\documentclass[sigconf, screen]{acmart}
\usepackage{booktabs,stfloats,microtype} 
\usepackage{enumitem}

\AtBeginDocument{%
  \providecommand\BibTeX{{%
    \normalfont B\kern-0.5em{\scshape i\kern-0.25em b}\kern-0.8em\TeX}}}

\copyrightyear{2022}
\acmYear{2022}
\setcopyright{acmcopyright}\acmConference[CHI '22]{CHI Conference on Human Factors in Computing Systems}{April 29-May 5, 2022}{New Orleans, LA, USA}
\acmBooktitle{CHI Conference on Human Factors in Computing Systems (CHI '22), April 29-May 5, 2022, New Orleans, LA, USA}
\acmPrice{15.00}
\acmDOI{10.1145/3491102.3501955}
\acmISBN{978-1-4503-9157-3/22/04}

\usepackage{hyperref,quoting}
\quotingsetup{vskip=0pt,font={itshape,raggedright},rightmargin=0pt}

\begin{document}

%%
%% The ``title'' command has an optional parameter,
%% allowing the author to define a ``short title'' to be used in page headers.
\title{The Unboxing Experience: Exploration and Design of Initial Interactions Between Children and Social Robots}

%%
%% The ``author'' command and its associated commands are used to define
%% ``authornote'' and ``authornotemark'' commands

\author{Christine P Lee, Bengisu Cagiltay, and Bilge Mutlu}
\affiliation{%
  \institution{Department of Computer Sciences, University of Wisconsin--Madison}
  \country{Madison, Wisconsin, USA}}
\email{{cplee5,bengisu,bilge}@cs.wisc.edu}

% \author{Bengisu Cagiltay}
% \affiliation{%
%   \institution{Department of Computer Sciences, University of Wisconsin--Madison}
%   \country{Madison, WI, USA}}
% \email{bengisu@cs.wisc.edu}

% \author{Bilge Mutlu}
% \affiliation{%
%   \institution{Department of Computer Sciences, University of Wisconsin--Madison}
%   \country{Madison, WI, USA}}
% \email{bilge@cs.wisc.edu}

\renewcommand{\shortauthors}{C. Lee, B. Cagiltay, \& B. Mutlu}

\begin{abstract}

% We posit that how robots are initially introduced to children significantly shapes their perceptions of and experience with the robot. 

% As social robots are increasingly introduced to children's lives as technological devices and companions, a key factor that shapes their perceptions of and interactions with robots is how they are first introduced to it. This initial experience, which has emerged in popular media as ``unboxing'' new technology, 

% are more frequently introduced into children's lives, the most appropriate manner in which they can be introduced (i.e., as a technological devices or companions) remains undefined. The way children experience this introduction and the impact it has on their perception and relationship with the robot also remain unknown. 
    
Social robots are increasingly introduced into children's lives as educational and social companions, yet little is known about how these products might best be introduced to their environments. The emergence of the ``unboxing'' phenomenon in media suggests that introduction is key to technology adoption where initial impressions are made. To better understand this phenomenon toward designing a positive unboxing experience in the context of social robots for children, we conducted three field studies with families of children aged 8 to 13: (1) an exploratory free-play activity ($n=12$); (2) a co-design session ($n=11$) that informed the development of a prototype box and a curated unboxing experience; and (3) a user study ($n=9$) that evaluated children's experiences. Our findings suggest the unboxing experience of social robots can be improved through the design of a creative aesthetic experience that engages the child socially to guide initial interactions and foster a positive child-robot relationship.

% The findings from the free-play and co-design studies informed the development of a prototype box and a curated unboxing experience, which were assessed in the third evaluation study. 

    %Meanwhile, the unboxing phenomenon is observed in day-to-day consumer products, particularly popular with children to make up a whole genre of video production.

  %Our thematic findings suggest that children appreciated the physical design of the box, enjoyed its social entity, and had a positive perception on the designed unboxing experience. 
\end{abstract}




\maketitle

\begin{figure}[t]
  \includegraphics[height=2.5in]{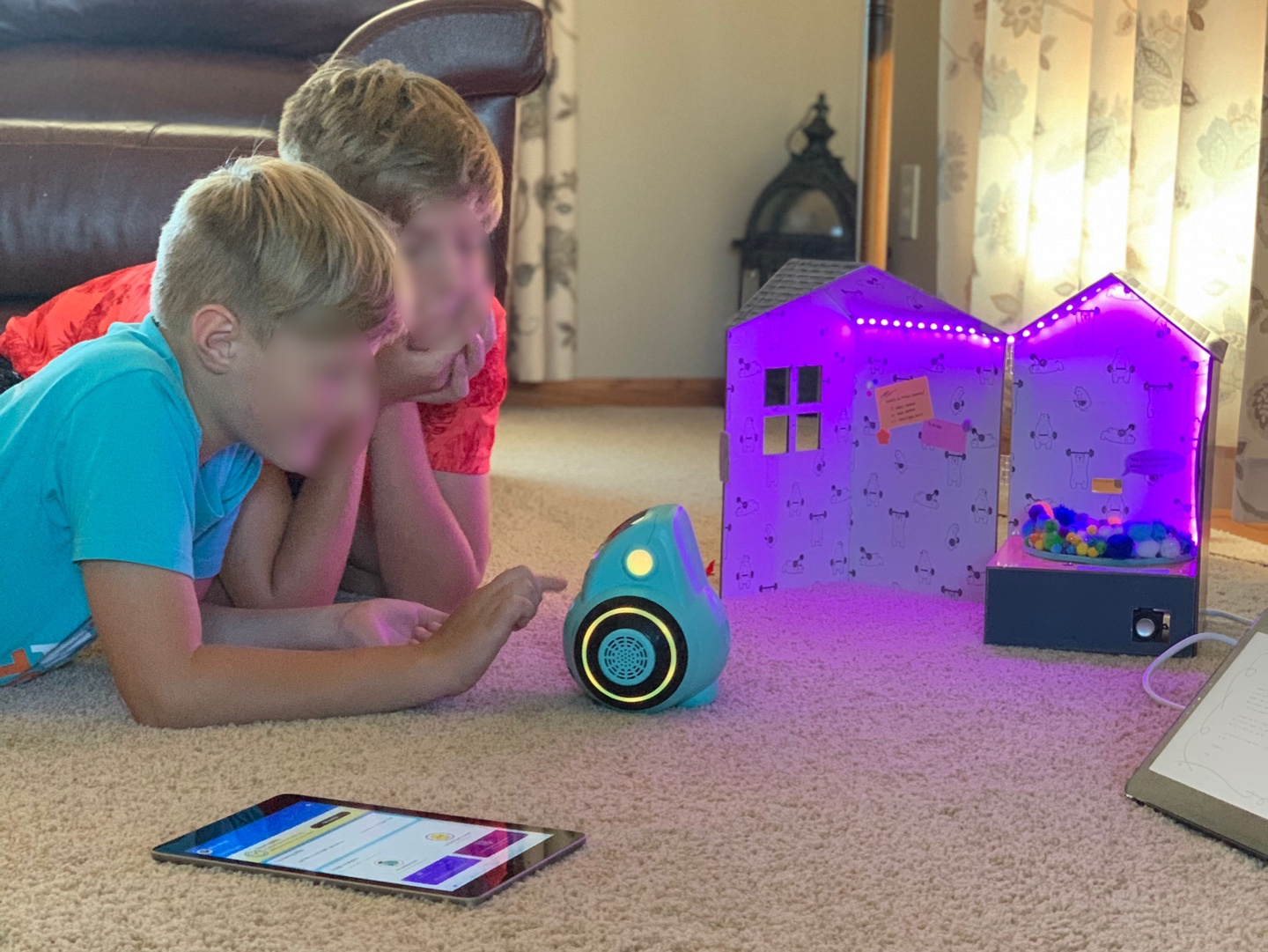}
  \caption{\textit{Children Interacting with Our Designed Unboxing Experience for a Social Robot ---} In this paper, we draw on the emerging phenomenon of ``unboxing'' consumer products to improve initial interactions between children and social robots. We explore the design space of unboxing social robots with children as co-designers, develop a curated unboxing experience, and evaluate our design. Our work explores and characterizes a new design space for improving user experience with social robots. }
  \label{fig:teasor}
%   \vspace{-6pt}
\end{figure}

\section{Introduction}
Social robots are emerging as the next generation of consumer products for day-to-day child environments, ranging in uses from entertainment to companionship and from education to healthcare \cite{kennedy16, Giannopulu12}. Such robot interaction with young users requires extra attention compared to other human-robot interactions due to the potential role robots might play in children's neurophysical, physical, and mental development. Fundamental to making child-robot interaction natural and integrated into their everyday lives is that the robot can successfully establish and maintain interest and rapport with the child. Although there is greater characterization and understanding of the design space for social robots in the human-robot interaction literature \cite{deng2018formalizing,deng2019embodiment}, this body of work focuses on task-based and nominal human-robot interactions, and not on the entire lifecycle of the use of social robot technologies in day-to-day settings, nor on the different stages of this lifecycle, including introduction, routine use, and abandonment. 

Prior research highlights the importance of the \textit{introduction} and \textit{initial interactions} with a social robot, as user experience at this stage substantially affects user perceptions and long-term interactions \cite{prusmann20}. Prior work has shown that a successful introduction process can serve to reduce uncertainty about the affordances of a robot and shape users' subsequent interactions \cite{fischer2011people}, while initial impressions of technology and design can influence user decisions, task performance, and user preferences \cite{schluter2021no, Kim16, cafaro16}. The ``social'' nature of robots further complicates these initial interactions, as it is not well understood whether these products should be introduced as companions with names and personalities; technological devices with features and functions; or both. 

A phenomenon related to initial interactions with technology and consumer products is the experience of ``unboxing'' \cite{craig17}, \textit{i.e.,} the process of opening the packaging of a product, taking the product out of its box, inspecting its parts, and starting use. This phenomenon is significant for experience design, not because of the mechanical action of taking the product out of its box, but because of the excitement and interest involved in ``meeting'' the product for the first time, appreciating its design, and exploring its capabilities. A large volume of unboxing videos capture the unboxing of a wide range of products, including children's toys, adult cosmetics, high-tech consumer devices, and so on. Between 2010 and 2014, the unboxing genre grew by 871\% and continues to be one of the most popular genres in online media \cite{kelly14}. While there are countless examples of unboxing experiences through online videos, little is known about how unboxing can impact children's experiences with social robots and the formation of child-robot relationships.

We argue that unboxing of a social robot, and initial interaction in general, is a critical stage of user experience and a unique opportunity to improve the long-term success of the design. Therefore, a better understanding of the process and factors involved in unboxing as well as knowledge of how these initial interactions shape user experience can inform the design of unboxing experiences that can facilitate the forming of child-robot relationships. The design space for supporting initial interactions through finer unboxing experiences presents remarkable potential that is yet untapped.

To address this gap in our understanding and explore the design space for unboxing, we conducted three studies that (1) examined baseline unboxing experiences; (2) engaged children in designing a new unboxing experience; and (3) evaluated the new unboxing experience (Figure \ref{fig:teasor}). Our findings indicate that a successful unboxing experience can enrich children's experience with social robots. Our work makes the following contributions:

\begin{enumerate}
    \item Identification of ``unboxing'' as a social phenomenon and a design space for user experience;
    \item A characterization of the process of unboxing a social robot product based on data from a field study;
    \item Co-design of new unboxing experiences for a social robot with children;
    \item A novel prototype box and a curated unboxing experience for a social robot;
    \item A set of design recommendations and guidelines for social robot designers;
    \item Empirical understanding of the impact of the unboxing experience on children's perceptions of social robots.
\end{enumerate}
%   \vspace{-12pt}

\section{Related Work}
Our work builds on prior literature on the recently emerged phenomenon of unboxing consumer products as well as research through design as an approach to designing new unboxing experiences. We also draw from prior work on child-robot interaction and studies that engaged children in the design process. The paragraphs below briefly summarize this body of work. 

\subsection{The Unboxing Phenomenon}

Unboxing is a rapidly emerging phenomenon in online videos and mainstream media \cite{craig17, kelly14}. These videos are not limited to particular types of products or directed at particular consumers, but they all share a focus on the momentary experience of unboxing. A number of related works explain the growing trend of unboxing videos within children on various consumer products including social robots \cite{Chesher19, marsh15}. The unboxing phenomenon has not only expanded the design space of products beyond their technical and functional aspects, but it has also encouraged manufacturers to consider product packaging, styling, and potential emotional responses from consumers \cite{kim18, desmet}. Manufacturers of technology products hold patents on product packaging and carry out studies on how the unboxing experience can evoke particular emotional responses from customers \cite{Heisler_2012}. This increasing interest in unboxing highlights the potential that unboxing holds to positively shape user experience and initial interactions with technological products.

\subsection{Research through Design in HCI}

\textit{Research through design} enables a broader form of inquiry for designers who aim to make the \textit{right thing}, as opposed to making the \textit{thing right}, through experimental thinking and exploratory design methods \cite{buxton07}. The approach of research through design allows for unlimited flexibility in solving unknown, complicated, or novel problems \cite{luria2021research, buch16}. This approach offers several benefits to the HCI community, including (1) new opportunities to reinterpret conventional understandings and identify gaps in extant models; (2) extended design boundaries through unlimited design choices and actions; and (3) novel perspectives and design methods for problematic situations \cite{zimmerman07}. Several studies have adopted a research-through-design approach to investigate design opportunities for child-robot interactions from a developmental perspective, introducing new design paradigms through collaboration with parents, educators, and children \cite{zaga17, zaga21}. 

\subsection{Children's Perceptions of Social Robots}

%\subsubsection{Appearance/ First Impression /Familiarity, first interaction}
Prior work has shown that children perceive social robots to be closer to human beings than mere machines \cite{breazeal16}. Research has suggested that the factors the underlie these perceptions include familiarity, appearance, first interactions, and impressions \cite{westlund17, hoeflich15}. 
First, children's familiarity with robots can contribute to a positive user experience with robots. Familiarity helps generate common ground and a positive atmosphere that facilitates the building of trust, enthusiasm, and receptivity toward the robot ~\cite{kanda04}. Previous work has sought to build familiarity by implementing introductory activities to acquaint children with a robot before experimental activities ~\cite{vogt17}. Other research has shown that appearance, first interaction, and first impressions (\textit{i.e.,} judgements made during the initial encounter based on limited information~\cite{wood13}) to be closely linked and to influence children's perceptions toward robots. Prior work has shown that appearance of the robot, such as facial expressions, non-verbal cues, and physical attributes, can impact children's likability, receptivity, and expectations toward robots ~\cite{barajas20, westlund17}. These studies have also highlighted the importance of entertaining and engaging first interactions, as these elements serve to ease the tension between humans and robots while improving users' perceptions of anthropomorphism and likability ~\cite{zlotowski15, prusmann20}.
Often heavily influenced by appearance and initial interactions, first impressions impact user engagement, expectations, perceptions, and behavior \cite{haring18, manja08, powers06}. These first impressions have been shown to impact the development of initial and long-term relationships~\cite{human13, sunnafrank04}, trust~\cite{xu18}, and user experience \cite{petrak19}. Research has also shown that initial bias or mental images formed during this time seldom recover \cite{prusmann20}. 

\subsection{Co-Design with Children}

Over an extended period, the research community has recognized the potential and capability of children as partners in designing interactive robot technology \cite{farber02, garz08,yip17, yip13}. The participatory design approach, which emphasized designing \textit{for} and \textit{with} children, has led to methodological innovations for equal, cooperative partnerships between children and adults when designing technology \cite{druin99, druin01}. The participatory approach foregrounds children's voices during the design process, thereby emphasizing and prioritizing communication with children to understand their needs, barriers they may face, and design gaps that have been overlooked in the current technology \cite{oliveira21}. By exploring potential solutions and future designs, co-design with children has been seen to successfully support the creation of robotic technologies \cite{metatla18, kaze17, metatla20, cagiltay2020investigating}. 

To extend the findings of previous works and begin investigating initial interactions between children and social robots, we aim to address the following research questions: 

\begin{itemize}
\item [] \textbf{RQ1:} What constitutes the unboxing experience of social robots for children?
\item [] \textbf{RQ2:} How does the unboxing experience of an in-home social robot influence a child's perceptions of the robot? 
\end{itemize}

\begin{figure*}[!t]
  \includegraphics[width=\textwidth]{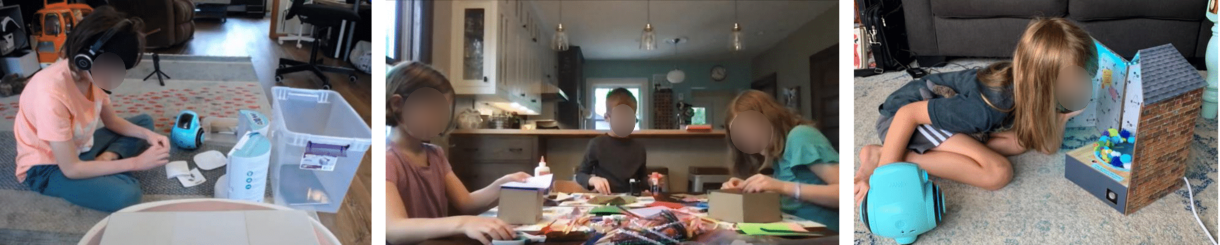}
%   \vspace{-12pt}
  \caption{\textit{Exploration, Design, and Evaluation of the Unboxing Experience ---} Our exploration and design of children's unboxing experiences involved observations of children's unboxing of a social robot (left) and co-design sessions to design new unboxing experiences (middle). Drawing on the ideas developed in the co-design sessions, we developed a curated unboxing experience that involved a ``social'' box and assessed how our design shaped children's unboxing experiences of a social robot (right). }
  \label{fig:study}
%   \vspace{-6pt}
\end{figure*}

\section{Method}
To address our research questions, we conducted three studies:\footnote{Study resources, such as interview questions used throughout the studies, are provided at \url{https://osf.io/4shc8/?view_only=6561e17dc96f4b3a8c0e2454ecac82ae}} (1) an observation of how children may unbox a social robot product in a naturalistic environment; (2) co-design sessions to ideate and prototype novel social robot unboxing experiences; and (3) an evaluation of the refined design of an unboxing experience generated in co-design sessions. The co-design sessions also included feedback sessions with co-designers for final designs. A summary of the studies is presented in Table \ref{studysummary}, and scenes from each study are shown in Figure \ref{fig:study}.

For Studies 1 and 3, we used an off-the-shelf social robot product designed for children, particularly for entertainment and informal education, named Miko.\footnote{\url{https://miko.ai}}  The robot platform integrates cooperative and interactive activities appropriate for child-robot interactions \cite{Ros11}, such as dancing, telling jokes, story telling, games, and short yoga sessions. Miko originally comes in a cylindrical box with a magnetic opening and supplemental equipment (\textit{i.e.,} charging cord and manuals) packed underneath the robot. Studies 2.1 and 2.2 did not require the use of Miko or any other robot. Study 3 utilizes the Miko robot while replacing the original box with our prototype. 

Studies 1 and 2 were conducted over Zoom\footnote{\url{https://zoom.us}} to accommodate COVID-19 restrictions. Study 3 was conducted by the first author visiting participants' homes following with COVID-19 safety protocols. All studies took place within participants' homes to provide a naturalistic environment and capture authentic interactions. The procedures for all studies were reviewed and approved by the University of Wisconsin--Madison Minimal Risk Research Institutional Review Board under protocol \#2021-0269. All participants were recruited from the greater Madison, Wisconsin area through mailing lists for the University of Wisconsin--Madison staff.

\begin{table*}[h!]
    \caption{\textit{Summary of Studies ---}
    Families from Study 1 (Explore) were asked to participate in the first part of Study 2.1 (Co-design). Families who participated in Study 1 and Study 2.1 were asked to participate in Study 2.2 (Feedback). New families were recruited for Study 3 (Evaluate).
    }
    \label{studysummary}
    % \centering
    \small
    \begin{tabular}{p{0.1\textwidth}p{0.2\textwidth}p{0.2\textwidth}p{0.2\textwidth}p{0.2\textwidth}}
         \toprule
  & \textbf{Study 1: Explore}& \textbf{Study 2.1: Co-Design}& \textbf{Study 2.2: Feedback}& \textbf{Study 3: Evaluate} \\
         \midrule
 \textbf{Goal}   & Explore and define the unboxing experience & Represent children's perspectives in design & Solicit feedback on prototype & Evaluate the final experience with new children \\
  \hline
  \textbf{Participants} &  9 families, 12 children & 8 families, 11 children & 4 families, 6 children & 7 families, 9 children \\
 \hline
 \textbf{Method} & Observation & \multicolumn{2}{c}{Participatory design} & Observation \\
 \hline
 \textbf{Results} &   Identified design elements and social factors of the unboxing experience & Developed the box prototype, story, and initial interaction with the social robot & Implemented co-designer feedback in the unboxing-experience final design &  Identified what design elements improve the unboxing experience \\
 
          \bottomrule
    \end{tabular}
    % \vspace{-12pt}
\end{table*}

\subsection{Data Analysis}

We conducted a Thematic Analysis (TA) based on the recorded interviews and design sessions to identify themes related to children’s experiences, preferences, and perceptions toward the social robot unboxing experience. The first author was familiarized with the data through facilitating the study sessions, and the second author reviewed the study recordings. The interviews were transcribed using Zoom cloud services and were manually reviewed by the authors to correct for errors. The interviews were coded following the guidelines developed by \citet{clarke2014thematic} and \citet{McDonald19}. The first two authors then reviewed the transcriptions and iteratively developed and applied codes until agreement was reached. The codes were then grouped into clusters representing themes emerging from our study data. After all candidate themes were discussed and reviewed, the final themes are reported as findings. When reporting our findings, we use the notation F\textit{i} to refer to families, C\textit{i} to refer to the main child participant, and S\textit{i} to refer to siblings, where \textit{i} indicates family ID number.

\subsection{Study 1: Exploring Children's Social Robot Unboxing Experiences}

\subsubsection{Study Design}
In this study, we aimed to explore and identify design factors from children’s experiences and preferences that support the unboxing experience and initial interactions with a social robot. The study lasted approximately 90 minutes and included two parts. During Part 1, children unboxed a Miko robot presented in its original packaging. We asked children not to turn the robot on, but rather observe the details of the packaging and its components (\textit{i.e.,} the exterior and interior of the box, manuals, and additional components). Part 2 of the study involved a free-play activity with the social robot, where children interacted with the robot by selecting from a set of available activities. The researchers provided limited guidance and responded to questions by children. Semi-structured interviews followed both parts of the study.

\subsubsection{Participants}
Nine families with at least one child (4 males, 5 females) aged 8--13 ($M = 10.3$, $SD = 1.2$) participated in Study 1. Children were predominantly identified by parents as white (88.89\%) or mixed (11.11\%). We refer to these families as F1--F9 and each eligible child as C1--C9. Two families (F4, F5) had siblings (S1, S2, S3) that did not meet the age range but were eager to participate and provide feedback (S1 and S2 for F4; S3 for F5). In total, twelve children (including siblings) from nine families participated. 

\subsubsection{Findings} 
In our analysis of data from Study 1, nine common themes emerged related to the unboxing experience of social robots: (1) the appearance of the box; (2) opening of the robot's box; (3) reusability of the physical box; (4) background story; (5) instruction manuals provided in the box; (6) robot greeting; (7) first interaction; (8) personality of the robot; and (9) shared activities with the robot. We grouped the first five themes under the meta-theme, ``physical design of the box,'' and the remaining four themes under the meta-theme, ``social interaction design.''

\paragraph{Physical Design of the Box} 
The five themes in this category are related to tangible, concrete, or aesthetic design factors, including (1) the appearance of the box, (2) opening of the robot's box, (3) reusability of the physical box, (4) background story, and (5) instruction manuals in the box. 

The first theme to emerge was the \textit{appearance of the box}, which includes descriptions, illustrations, and decorations on the exterior and interior design of the box. To children, these appearances were seen as a form of communication that shaped the first impression of the social robot. For example, \textit{C3:``Well, the box does matter because it tells what the robot can do about helping on education and fun.''} In support of the previous claim, children shared that a box should reflect or showcase what is inside. As a solution, children (C5, C6, C7) suggested that more detailed decorations can make the box to be more informative and interesting. When asked how the box should be decorated, several ideas were suggested: a window to provide a glimpse of the robot (C6, C7); a hole for a charging cable in the box (\textit{i.e.,} a way for the robot to ``eat'') (C3); customizable parts for decorations (\textit{e.g.,} drawings, stickers, notepads) (C1, C6, C7, S1, C5, S3); illustrations and descriptions of the robot’s capabilities (C5, C6, C7); and wheels for mobility (S3). These suggestions highlight the importance of the visual design of the box in shaping children's experiences of unboxing a social robot and the richness of the design space for supporting the unboxing experience. 

%box
Children also mentioned the importance of an \textit{easily openable box}. During the free play session, one child (C3) suggested adding a red arrow to more clearly identify the opening of the box after initially being unable to find it. One parent (F3) appreciated that the robot was not connected to the box with twist ties, as done with many conventional children's toys and products. The simple packaging lightened the role of parental assistance and allowed the child and the robot to directly interact from the beginning. All children emphasized that easy access to the robot is important because the longer it takes to start the interaction, the more tedious the unboxing experience feels and the less interested they become. 

%reusability 
\textit{Reusability of the box} was another theme that emerged, and children highlighted personal experiences related to this theme. One child (C7) suggested that boxes for electronic devices usually needed to be replaced, as the original packaging is usually not sufficiently sturdy to reuse or interesting to keep. This child described his hobby of building houses for robots the child owned, stating \textit{``oh yeah sometimes I take out some magnet tiles or cardboard and I make like a house for our robot Cosmo.''} Another child (C3) explained having a designated spot on a shelf for safe keeping of a robot the child owned. While many boxes focus on the temporary role of protecting the robot from damage, children expressed a desire for the box to have a permanent role of giving the robot a place to stay. 

%Children also mentioned that the relevance of the box to the robot affected their decision to reuse it. They explained that an exterior design that supported a background  as if the box had a shape more interesting or relevant to the robot (i.e., house or spaceship), children would want to reuse the box. (S1, C3, C6, C7).
%as a more interesting box was more likely to be seen as reusable appearance makes it seem like there is a story
%bg stroy -> reusability, apprearance -> bg stroy
%When asked why the would pick the box shape to be a house, \textit{``\textbf{C7}: it(Miko) feels like it kind of needs a house, it looks kind of lonely.''}
%bg story

The \textit{background story} of the robot was another notable theme that children enthusiastically discussed (C4, S1, C5, S3, C3, C6, C7). Some enjoyed generating new stories about where the robot came from and why, while others added short introductions that explained the robot's capabilities. Children expressed that the robot sharing its background story during the initial introduction would pique their interest and create a stronger sense of connection. Specifically, one child (C7) said, \textit{``Yeah I mean [the background story] kind of gives me an idea of what [Miko] is capable of. And if I knew where [Miko] came from it [would] make me feel a bit more connected with it.''} Moreover, the following conversation suggests potential presentation methods and further highlights the importance of a background story for the robot:

% \begin{itemize}
\begin{quoting}

\textbf{C7}: I might have like kind of a weird idea, but I might like put the robot on the box and have the screen or something kind of reading a story. It [would] kind of be a good design and functionality. %in a good area

\textbf{Experimenter}: Why would you choose the screen?

\textbf{C7}: It'd be more interesting than putting all the text on the back of the box or something, it would kind of fit everything together and might show what the robot can do.
% \end{itemize} 
\end{quoting}

Related to the previous theme of reusability, children (C3, C6, C7, S3) explained that the box would be more useful and realistic if it aligned with the robot's story. Based on an activity one child (C3) completed with the robot, the child thought that Miko was from outer space and suggested that the cylindrical box should look and act as a spaceship for the robot. C3 explained, \textit{``the shape of the box can also fit the backstory of the box. Like from outer space or something, so it has this [cylinder] container. So when I'm done with the robot, I would put it back in ... like sit in a corner or something until when I want to use it again. It can be a station for it, like a house.''} These creative stories allowed children to build a stronger interest in and connection with the robot by providing additional content that children can relate to and that can help children form common ground with the robot. 
%story is important for the box

%instruction manual
The final theme we identified related to physical design was the presentation of the \textit{instruction manual}. Children (C1, C6, C7) expressed that the robot's \textit{instruction manual} was complicated and cumbersome to read, but the quick start guide was helpful during the free-play activity. While the instruction manual is necessary to provide a detailed explanation of the robot's operation and capabilities, the quick start guide is meant to provide a brief overview and can therefore be delivered in various ways. Children (C3, C5, S3, C7) explained that instructions in the form of comic books, videos, or stories would encourage engagement and attention compared to verbose instruction manuals. Children's suggestions indicate that novel and creative ways of delivering instructions and guides can create more engagement and interest in the robot and the activities facilitated by the robot.

\paragraph{Social Interaction Design}
The four themes in this category are focused on the relationship built between children and social robots, including (6) first interaction, (7) robot greeting, (8) personality of the robot, and (9) shared activities with the robot. %(10) Delivery of the robot.

%personality, greeting, first interaction, shared activities 
Children emphasized the importance of the \textit{first interaction} with the social robot as a significant moment where children decide on whether or not maintain interest in and curiosity about the robot. Some children expressed confusion or boredom during the setup process (\textit{i.e.,} connecting to the WiFi, charging, and updating), as it led to delayed playtime. Children (C4, C5, C7) explained that a delayed playtime quelled their interest and illustrated the need for a robot that was immediately ready to play. For example, one child expressed \textit{C7:``The WiFi setup was a bit frustrating. I wish I could just play with it.''} These comments open the possibility of the setup process being integrated into the initial interaction in the form of a social interaction between the robot and the child or a parent. 

Additional themes for the social interaction design of the robot were its \textit{greetings} and \textit{personality}. Children (C1, C4, S1, C7) felt that the robot's facial expressions, particularly the robot's eyes, determined its personality. One element of the robot's facial expression was the variation in the shape of its eyes (\textit{e.g.,} hearts, sunglasses, happy eyes, crying eyes), which the children interpreted as the robot's feelings ranging from excitement to sadness. C4 stated, \textit{``I liked how the eyes changed. The heart eye one would be good. Oh, I liked the sunglasses one too. [Miko] could walk by and say `hi' with his sunglasses on. It makes the robot look fun to play with.''} Another child (C5) suggested that the robot share its name and ask for the child's name during the greeting, which would help maintain the child's interest and engagement. Children expressed that the initial greeting and personality of the robot can greatly influence their first impression of the robot. 

Following the initial greeting, another common theme we identified was a preference for \textit{shared activities} with the robot. Children often asked the robot to dance and play songs, and some joined the robot in dancing. Children with siblings explained that they most enjoyed the dancing activities because they could participate as a group. Some children also expressed that being able to ask questions through voice recognition was entertaining and made the robot feel realistic. However, some children occasionally felt discouraged by the mostly one-way conversation. Additionally, one child (C6) suggested that the robot uses icebreaker activities during the initial interaction, stating \textit{C6:``So I think that Miko should have a little, different types of icebreakers. Like maybe a little a backstory about you or it can be like a little game to get to know each other.''} 

\subsection{Study 2.1: Co-Designing Social Robot Unboxing Experiences}
\subsubsection{Study Design}
Study 2 included participatory design with children to understand their perspectives regarding and expectations of the social robot unboxing experience. This study was anchored by the nine design themes identified from Study 1. In this study, children designed their own social robot unboxing experience, which included the physical design of the box and social interaction design. The study lasted approximately two hours with three parts: (1) a sorting activity of the previously defined design themes, (2) a box design activity, and (3) a social robot interaction design activity. Semi-structured interviews followed each phase.

\begin{table*}[!t]
    \caption{\textit{Design Prompts} --- Examples of the design prompts used in Study 2.1.}
    \label{prompts}
    % \centering
    \small
    \begin{tabular}{p{0.35\textwidth}p{0.6\textwidth}}
         \toprule
 \textbf{Design Components} & \textbf{Design Prompts} \\
         \midrule
 \textit{The Appearance of the Box}   & If you are the designer, how would you design the appearance of the \\
  & box for a social robot? What type of features/parts would it have? \\
  \hline
  \textit{Reusability of the Physical Box} &  If you are the designer, would you design the reusability \\
 & of the box for a social robot? If so, what would the box be reused for?\\
 \hline
 \textit{Background Story} & If you are the designer, would there be a background story\\
 & for the social robot? If so, what would the story be?\\
 \hline
 \textit{First Interactions} &   If you are the designer, how would you design the social robot's first \\
  & interaction? Who would make the first move? How would it be done? \\
 \hline
 \textit{Shared Activities with the Robot} &  If you are the designer, what activities would you want the robot to have? \\
 & What would they look like? Who would be involved? \\
          \bottomrule
    \end{tabular}
    % \vspace{-12pt}
\end{table*}

\textit{(1) Sorting Activity of Design Components: } During the first activity, children sorted nine envelopes, each labeled with a design theme from the set of themes identified in Study 1, based on what they perceived to be the most to the least important. The purpose of the sorting activity was to facilitate discussion of which design points were most important to the co-designers and why.%, also to reveal other potential design points.

\begin{figure*}[!b]
  \includegraphics[width=\textwidth]{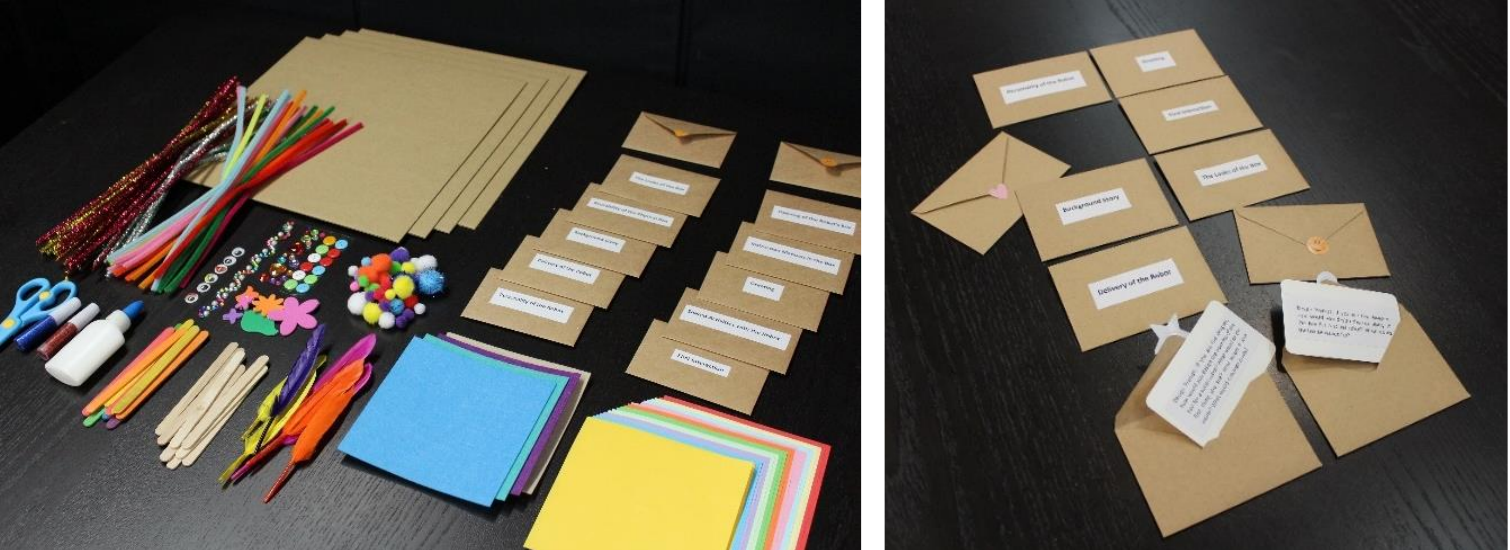}
%   \vspace{-12pt}
  \caption{\textit{Resources Used in Co-design Sessions ---} 
  As co-designers, children contributed to the design of a new unboxing experience by developing a narrative, designing a box that reflected the narrative, and envisioning initial interactions that children may have with their own social robot. Resources were provided for ideation (left) and fabrication (right).}
  \label{fig:materials}
%   \vspace{-6pt}
\end{figure*}

\textit{(2) Box Design Activity: } Before starting the box design activity, researchers asked children to separate a pile of five envelopes related to the physical design of the box, including (1) the appearance of the box, (2) opening of the robot's box, (3) reusability of the physical box, (4) background story, and (5) instruction manuals in the box. Each envelope labeled with a design theme contained a prompt that gave a minimal explanation of the labeled theme (\textit{e.g.,} Table \ref{prompts}). Children were also provided with an identical design challenge (example presented below) to serve as an objective for their design activity. After reading the design prompt and Design Challenge 1, children used the provided resources shown in Figure \ref{fig:materials} (\textit{i.e.,} arts-and-crafts materials, cardboard sheets, scissors, glue) to construct their designs. 

\textit{\textbf{Design Challenge 1:} Design a Box and an Unboxing Experience\\
Now, imagine you are a designer who works for a robot company, and your family members are your coworkers. Using the materials that are in the bags of supplies, you are in charge of designing the robot's box, its delivery, and unboxing. Try to design a prototype box and then describe what the delivery and unboxing process will look like.}

\textit{(3) Social Robot Interaction Design Activity: } The last activity focused on designing a social robot, its social behaviors, and its first interactions with its user. This activity used the remaining four envelopes related to social interaction design, including (6) robot greeting, (7) first interaction, (8) personality of the robot, and (9) shared activities with the robot. After reading the design prompts and Design Challenges 2 and 3, provided below, children freely designed social robot figurines with the provided resources. Finally, children designed and presented their envisioned social robot interactions using the figurine through role-play or description. 

\textit{\textbf{Design Challenge 2:} Design a Social Robot Figurine \newline
Design or draw a social robot to help us with the next tasks. Once you finish designing, you can stick your robot on one or more Popsicle sticks, imagining that it is a puppet or your robot prototype.}

\textit{\textbf{Design Challenge 3:} Role-play with Your Social Robot Figurine\newline 
Now imagine you, the designer, and your coworkers are preparing for a presentation to potential investors or customers. Your goal is to describe or act out how the robot's first interaction with its user will look like. You can use the materials you have in the bags of supplies, the box you designed, and your social robot figurine.}

\subsubsection{Participants}
Eight families with at least one child (4 males, 4 females) between ages 8--13 ($M = 10.5$, $SD = 0.92$) participated in Study 2.1. Children were predominantly identified by parents as white (87.5\%) or mixed (12.5\%). Seven participant families from Study 1 also participated in Study 2.1 (F1--F7), while two (F8 and F9) did not express interest in continuing participation. We recruited one additional family (F10) for Study 2.1. In total, 11 children (including siblings) from eight families attended this study. 
 
\subsubsection{Findings}

\begin{figure*}[!b]
  \includegraphics[width=\textwidth]{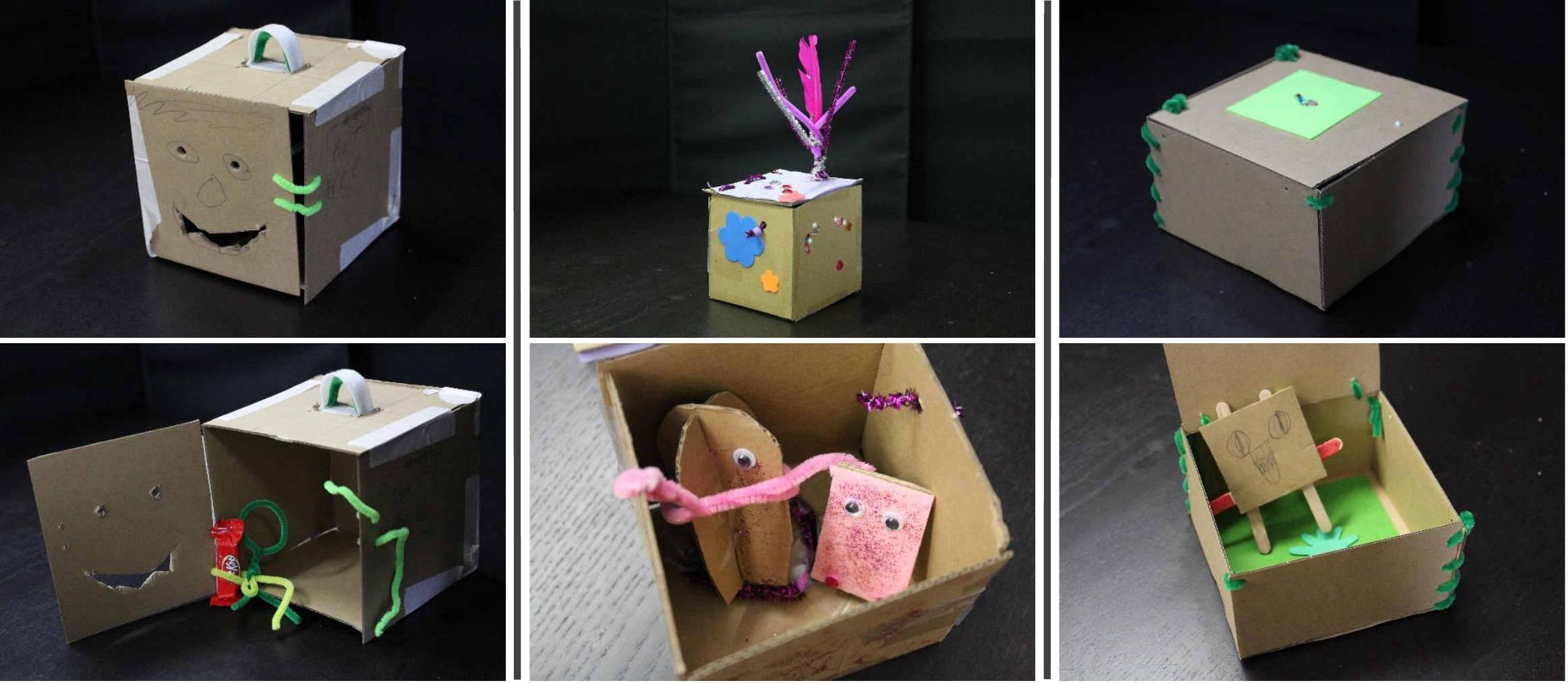}
  \caption{\textit{Prototypes from Co-design Sessions ---} The ideas generated in the co-design sessions inspired our final design of the unboxing experience. The box on the (left) presents the idea that the box itself can serve as a ``social entity'' to support the social robot. The box in the (middle) illustrates the idea of the box as a home with creative decorations to make the robot and box seem ``friendly.'' The box on the (right) represents the idea that the design of the box can reflect a theme that matches the robot's characteristics, \textit{e.g.}, the ``active'' robot lives within its ``jungle'' themed box. 
  }
  \label{fig:PDbox}
%   \vspace{-6pt}
\end{figure*}

\paragraph{Designs from Co-Designers}
We present the designs generated by children, which includes the design of a social robot, its box, and the initial interactions between children and the social robot. While our analysis included work by all children, we present the work of four co-designers whose designs made the strongest and most concrete contributions to our findings. Multiple designs are shown in Figure \ref{fig:PDbox}.

\textbf{C3} designed a cylinder-shaped box that resembled a house with a window, a roof, and a door with a doorknob for easy opening. C3 explained that the box, or house, was to be a place that \textit{``the robot can come back to rest after playing outside''} with the child. C3 also created a charging hole on the wall of the house for the robot to be charged from the inside. C3 explained that this allows the box to be reusable, by giving the robot a place to \textit{``stay, sleep, and eat.''} C3 arrived at this design because the social robot felt more like a friend than a regular toy and thus wanted to give it a designated place to \textit{``sleep over.''} Additionally, C3 explained how the robot should ask the child's name during the initial greeting. This gesture by the robot would make the child feel \textit{``happy,''} as the robot would be expressing interest in getting to know the child. C3 also described that the first interactions need to involve interactive, simple, and easy activities, such as Q\&A sessions, racing, or ball games. 

\textbf{C7} designed a story for the box and the robot, explaining \textit{``the monster box ate the robot and lets it live inside.''} C7 described that the box and the robot were interdependent and interactive, each having a part of the whole narrative. The monster box had a role of protecting the robot from danger, while the robot lived inside sharing food with the box. C7 also carved a smiley face in the front of the box to give it \textit{``a happy and nice personality.''} C7 emphasized that facial expressions can give the box some personality and make it into something more than ordinary packaging, describing that the idea originated from the logo on Amazon\footnote{\url{http://amazon.com}} delivery boxes. In C7's design, the first greeting involved exchanging names, and the first interactions involved \textit{``[the robot] showing off some abilities through activities like tutorials.''} C7 explained that this type of greeting enables the robot to show its capabilities and give children an idea of what to expect during the interaction. C7 claimed that it is important for the robot to show its capabilities during the introduction for the user to not over or underestimate the robot. 

\textbf{C2} made a box shaped like a car for a racing robot. This box was not only meant be a place for the robot to stay, but it was also a toy for the child to reuse and play along with the robot. C2 explained that when people were not looking, the robot drives the car itself and plays together. When describing the design, C2 mentioned that it is important for the robot and box to have a related theme to prevent confusion and enjoyed the idea of the box serving as an additional toy beyond the robot itself. C2 also shared how the background story would be \textit{``[the robot having] a rusty car [repaired] into a new, shiny race car to race in.''} In C2's design, the instruction manual appeared on the screen of the car. C2 expressed preference towards a screen over printed instruction manuals, as it makes it more fun and easier to follow. C2 described the first activity to be the robot asking the child to join a race game, while the robot would race in its car-shaped box. 

\textbf{C4} designed a box with a jungle theme for an ``active'' robot. The exterior and opening of the box was secured with vines, while the interior was designed to look like a grass field. While describing the idea of reusing the box to keep the robot inside, C4 mentioned that the jungle box gives the robot a place to run around and sleep. C4 also mentioned how the robot should be delivered by an elephant, to support the story of the robot coming from the jungle. C4 emphasized that the first interaction activities need to be entertaining enough to capture children's interest in the robot. C4 specifically suggested a dance activity for the first interaction, as it was \textit{``the most fun activity to do with a robot''} and could be played as a group. A sibling of C4, S1 called the box ``the treasure box,'' focusing on personalized decorations. The handle of the box was colorfully decorated with colorful materials to make it easy to open the door to the robot's house. S1 explained that personalized decorations made the child feel more attached to the box. The box contained a stool for the robot to stand on, with the robot's eyes looking toward the user. S1 explained that this position of the robot was intended for the child to make eye contact with the robot first when unboxing.

\paragraph{Designing the Unboxing Experience} 
Building on the ideas that were developed in the co-design sessions, we designed a novel unboxing experience with a four-phase interaction and a unique box for a social robot. 

Our interaction design for the unboxing experience of social robots involved the following phases: (1) prior interaction, (2) packaging, (3) first interaction, and (4) first impression. The \textit{``prior interaction''} phase aimed to familiarize children with robots. The resulting familiarity would help children build connections and understand what to expect before meeting the robot. The \textit{``packaging''} phased aimed to amplify the aesthetic potential of the robot box design, for packaging to go beyond the conventional role of protecting its contents and to support the child's initial interaction with the robot. The \textit{``first interaction''} phase focused on building a social connection between the child and robot through interactive activities that were engaging, interesting, and comfortable for the child. In the \textit{``first impression''} phase, the interaction focuses on the child forming positive judgments of the robot based on interactions in the previous phases, potentially increasing receptivity toward and rapport with the social robot. 

We designed a new box for the robot that had its own social entity to support the interaction between the robot and the child. The box was designed to help the robot, acting as a ``bridge'' to connect the robot and the child. In this setting, the box would contribute to building a social atmosphere to improve the child's experience. Our goal was to give the box a social presence by introducing it as a \textit{``home and butler for the social robot.''} The box was envisioned to provide verbal instructions for the introductory activities and assistance to questions regarding the robot. The box communicated through social cues, including lights and audio, in each activity. Moreover, the interior design of the box (\textit{e.g.,} wallpaper and pockets for instruction manuals) reflected the robot’s purpose of being with the child. For example, the wallpaper would be decorated with books for a robot designed to encourage reading. 

\subsection{Study 2.2: Feedback for the Designed Prototype}
\subsubsection{Study Design}
Before finalizing the designed unboxing experience, we discussed our prototype with the co-designers for feedback. The feedback study lasted approximately thirty minutes and was conducted remotely over Zoom. During the study, the researchers presented our curated unboxing experience to the children, which included the social robot sending a letter to the child as prior interaction, a designed box with a social entity, and interactive activities as the social robot’s first interaction. The first prototype box presented to the children was shaped like a house with windows, a magnetic opening, and a hole in the wall for charging. The exterior of the box was plain with no added texture, while the interior had wallpaper with books to support the social robot's role of encouraging children to read. The box was given a social character as a butler to assist and support the interactions between the child and the robot. As a butler, the box would explain the first interaction activities and answer questions about the robot to children. As light and audio components were not yet implemented within the box, researchers explained the envisioned design to children through illustrations. The presentation of the design was followed by open-ended discussions with co-designers about the overall unboxing experience. 

\subsubsection{Participants}
We recruited families to participate from those that attended Study 2.1. Four families (2 male, 2 female; F1, F3, F4, F7) with at least one child aged 8--13 ($M = 11$, $SD = 0.81$) participated in the feedback study. All children were identified by parents as white (100\%). In total, six children (including siblings) from four families attended the study. 

\subsubsection{Findings}
Feedback from children focused on improving (1) the appearance of the box and (2) the social aspects of the box.

% \paragraph{(1) Appearance/Aesthetic}
\paragraph{Appearance of the Box}
During the discussion, all co-designers were supportive of decorating the box. Children (C1, C4) suggested the plain exterior be replaced by a pattern that closely resembled that of a house, which would make the box and the narrative of the robot living inside feel more realistic. They explained that colorful or meaningful decorations (\textit{e.g.,} house designs, illustrations, robot descriptions, and name tags) would increase the child’s interest and curiosity when first meeting the robot. From the feedback, we added a wallpaper with prints resembling bricks, a roof design, and doors to the exterior of the box. 

The co-designers also encouraged designing the interior of the box, as interior design is often overlooked in conventional product packaging. All children were positive toward the wallpaper and pockets for manuals reflecting the robot’s theme, describing that this design would provide the children with clarity regarding the purpose of the robot. Moreover, one child (C7) specifically suggested to have multiple themes, as \textit{``some children may not like to read.''} As a result, we decided to include a fitness encouraging theme in addition to the reading encouraging theme for the social robot. The interior design of the box and activities with the robot would differ depending on the theme. 

Finally, feedback from the children indicated that other components should be placed inside the house. One child (C4) suggested that the robot's stool should be turned into a bed or a chair to fit the house design, thus we constructed comfortable-looking furniture for the robot to sit on. 

\paragraph{Social Aspects of the Box}

All children were supportive of the \textit{box having a social entity}, describing this as a clever way to reuse the box, avoid the conventional presentation of instruction manuals, and potentially bring the child and robot closer together. One child (C1) especially liked the box being \textit{``Freddy the butler,''} explaining the robot introduction and first interaction activities from a third person perspective. The child expressed that this form of instruction would improve the flow of the activities to be less confusing compared to the robot having to explain and perform the activities itself.

\begin{figure*}[!b]
  \includegraphics[width=\textwidth]{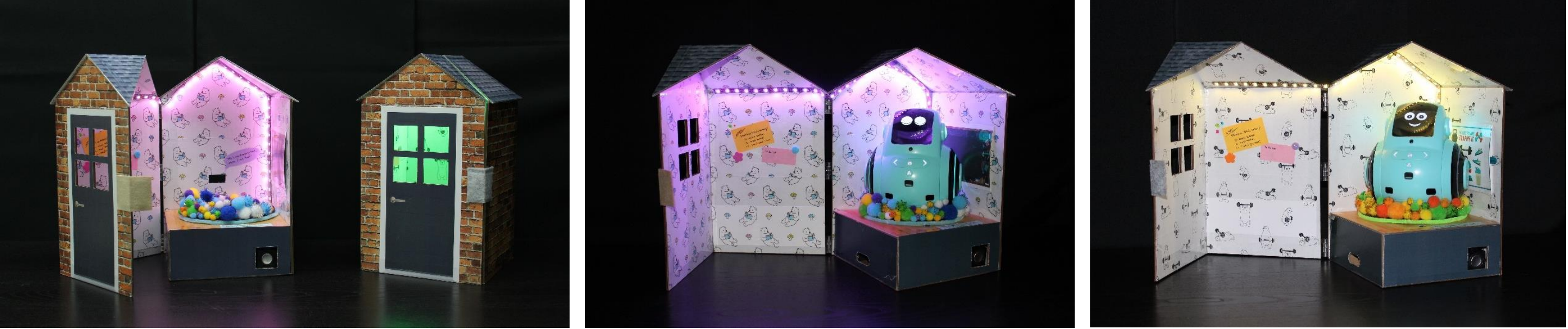}
  \caption{\textit{Freddy, the Butler ---} 
  The curated unboxing experience involved a box with a social entity called ``Freddy.'' Freddy served as a butler for the social robot Miko, with the purpose of facilitating interaction between the child and the robot. Freddy (left) was designed to be a home to the social robot Miko (middle and right). When first introduced to children, Freddy expressed verbal and non-verbal cues to introduce Miko and interactive activities which children can engage in with the robot. The box design included two versions: a \textit{reading} theme (middle) and a \textit{fitness} theme (right). Based on the theme selected, children engaged in different activities with Miko.
  }
  \label{fig:designedbox}
%   \vspace{-6pt}
\end{figure*}

The use of \textit{visual and audio cues} by the box especially received favorable feedback from all co-designers. Children explained that the lights and speech from the box was a simple and easy way to socially express and communicate, which would solve any confusion children may have during the first interaction. One designer (C7) also pointed out that the light and audio components were a good way to give the box a personality. 

Additionally, co-designers responded favorably toward \textit{background stories.} All children liked the idea of the robot having a background story for itself as well as the box having a story connected to the robot. Children mentioned that providing extra context on the robot’s identity, purpose, and its box would make the robot realistic and help children understand and relate. While all children agreed that introducing the background story during prior interaction would be appropriate, different ideas emerged on how the story should be delivered among children. One child (C1) suggested that a letter would be a simple but effective way to feel connected, particularly if the letter was specifically addressed to the child using the child's name. Another child (C7) suggested that the background story should be introduced by the robot or a third-person through the robot's speech to make the envisioned prior interaction accessible to children who do not yet read. Our final design adopted the idea of the robot sharing a letter addressed to the child explaining its backstory and introducing the box.

Although the idea of integrating a social entity in the box was supported by all co-designers, one child (C7) suggested that the box with social aspects should not overshadow the social robot itself. C7 described that it is important for the box to keep its boundaries, stating \textit{``I think that if you put enough stuff in it, the box kind of turns in to a robot itself and kind of defeats the purpose of a box.''} C7 also provided design suggestions for the visual cues of the box by explaining \textit{``I think the lights and stuff are a good idea that wouldn't be too overkill or anything. [The box] could also have a sensor or something that detects when the door opens or something.''} Based on this feedback, we added sensors to detect when the child would open the box and to automatically start the light and audio interactions. 

\subsection{Study 3: Evaluation of the Designed Unboxing Experience}
\subsubsection{The Designed Unboxing Experience}

We finalized our design of the social-robot unboxing experience based on the findings from the exploration, co-design, and feedback studies. Our curated unboxing experience involved a box with a social entity, a four-phase interaction design, and multiple themes for the social robot. 

\paragraph{Box: Freddy the Butler}
As shown in Figure \ref{fig:designedbox}, we designed our final box to have a social entity, integrating the character ``Freddy the butler.'' We used an Arduino\footnote{\url{https://www.arduino.cc}} microcontroller to simulate the social elements of the box, mounting all the components into the floor of the box. We implemented LED strips for lights, ultrasonic sensors for motion detection, speakers for audio responses, and buttons for control. The components of the box represent Freddy's social characteristics, as Freddy communicated and interacted with children using lights and audio cues. Although lights were not tied to specific emotions, lights welcomed children and supported activities during the first interactions. The appearance of the lights varied in each activity, such as blinking or changing color. Introduction of the robot and instructions for the activities were also delivered through audio using Freddy's voice. Thus, it was Freddy's job to guide, assist, and support children throughout the interaction activities with the robot. 
Freddy's exterior was designed to look like a house, while the interior was designed to reflect the robot’s theme (\textit{i.e.,} reading or fitness). The design of the house included a window, magnetic opening, a comfortable-looking stool for the robot, pockets for a quick start guide, sticky notes on the wall for button descriptions, and a charging hole. 

\begin{figure*}[!b]
  \includegraphics[width=\textwidth]{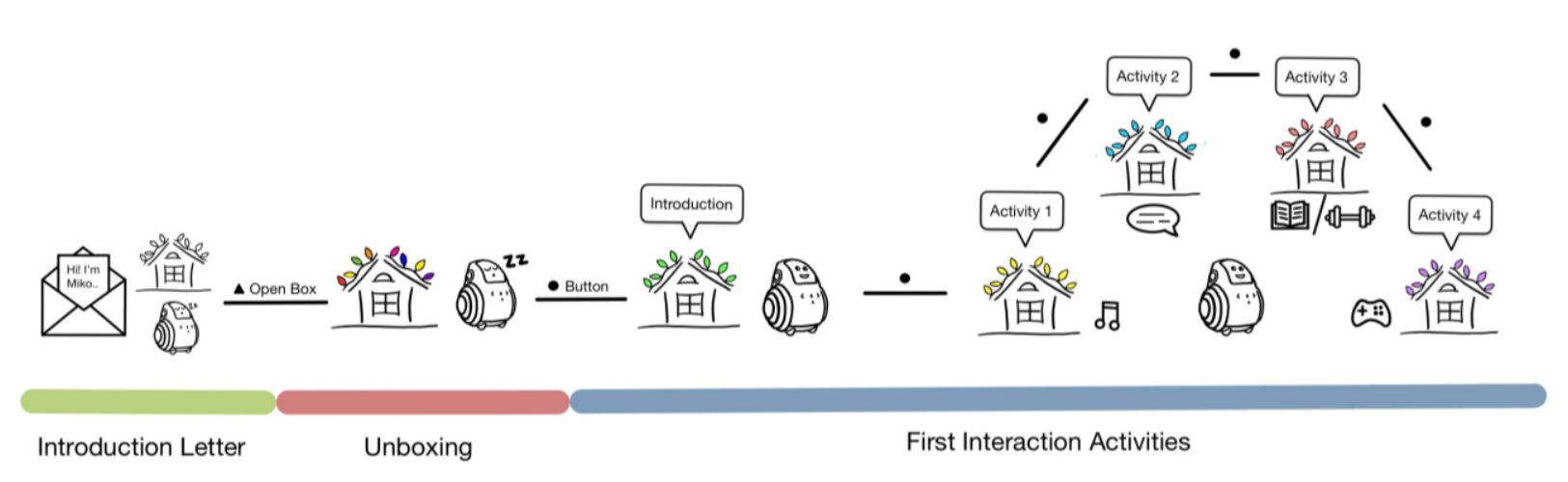}
%   \vspace{-12pt}
  \caption{\textit{Interaction Flow for the Designed Unboxing Experience ---} 
  The curated unboxing experience starts when children receives an introductory letter written by the robot. The letter is addressed to the child, containing a brief introduction of the robot and its box, ``Freddy the butler.'' The child then opens the box, and Freddy welcomes the child with blinking lights. The notes placed on the box guide the child to press a button for Freddy to start the audio. In the first audio speech, Freddy introduces the robot and explains its role as an assistant to the robot. After subsequent presses on the button, Freddy provides instructions for an interactive activity to play with the robot and changes its lights to various colors. The child engages in four different activities with Freddy and the robot, including dancing, telling jokes, storytelling, and playing games.}
  \label{fig:study3}
%   \vspace{-6pt}
\end{figure*}

\paragraph{Four-Phase Interaction Procedure}
The unboxing experience involved four phases---(1) prior interaction, (2) packaging, (3) first interaction, and (4) first impression---each including designed features. The full interaction procedure between the child and the social robot is shown in Figure \ref{fig:study3}.

In the prior interaction phase, we asked children to read a letter at the beginning of the unboxing experience. The letter was addressed directly to the child by the robot, including the robot introducing itself and its purpose, along with a brief introduction of its butler Freddy (\textit{i.e.,} the box). The letter concluded with the robot saying, \textit{``I am currently asleep right now, but Freddy will tell you how to wake me up!''} Children then met Freddy the butler and started unboxing. Once Freddy's door was opened, and this was detected by the sensor, lights blinked to welcome the child. Freddy's interior contained simple text instructions (\textit{e.g.,} \textit{``Press me! I will share some activities.''}) to guide children to press a single button. When the child first pressed the button, the lights changed to a static color, and Freddy played a pre-recorded audio file that briefly introduced the robot. Freddy then guided the child on how to wake and prepare the robot for activities. When the child pressed the button again, Freddy played another pre-recorded audio file with instructions on the first activity. The child engaged in the activity, then repeated this procedure four times for each activity. As the button was subsequently pressed, Freddy changed its light colors. 

\paragraph{Multiple Themes: Reading or Fitness}
The designed unboxing experience applied to two types of robot themes: a robot that encouraged \textit{reading} and another that encouraged \textit{fitness}. The interior design of the box, in its wallpaper textures and the design of the pockets, and the types of first interaction activities followed these two themes. The reading theme included the following activities: dancing, telling jokes, story telling, and playing games. The exercise theme included the same activities as the reading theme, except a yoga session replaced the storytelling activity.

\subsubsection{Study Design}
In the final study, we evaluated our designed unboxing experience by deploying it with newly recruited children. The study lasted approximately one hour and took place in the participants' homes. The study used the Miko robot, replacing Miko's original packaging with our prototype box. The study was structured into two parts. In Part 1, the first author set up the designed unboxing experience (\textit{i.e.,} putting the robot to ``sleep'' inside Freddy). Children then entered the study area and started the unboxing experience. During Part 2, children freely participated in activities with the robot. 

\subsubsection{Participants}
Seven families with at least one child (5 males, 2 females) aged 8--13 ($M = 10.4$, $SD = 1.4$) participated in Study 3. Children were identified by parents as being white (100\%). None of these families had experience with the Miko robot, nor had they participated in any prior design phases of this study. We report and refer to these families as F11--F17 and each eligible child as C11--C17. Two families (F12, F14) had siblings (S4, S5) %(C5, C7, C8) 
that did not meet the age range but were eager to participate and provide feedback (S4 for F12; S5 for F15).
% that did not meet the age range but was eager to participate and provide feedback, and F14 had two children eligible however we refer to the eligible sibling as S5. %(S4 for F12, and S4 for F14). 
Overall, a total of nine children from seven families attended this study. Families were provided with two available theme options (reading or fitness) prior to the visit. F12 and F17 preferred the reading robot, while the remaining families preferred the fitness robot. 

\subsubsection{Findings}
We present salient design topics that emerged from our data under two categories: (1) the appearance of the box and (2) social aspects within the unboxing experience.
%We show children's evaluation on the design topics within each category, followed by  how these features affected the child's perception. 
%weird
%We identified three main primary factors of the designed unboxing experience from the thematic analysis for study 3:
%(1) The Appearance/Aesthetic of the box %pom pom, wallpaper, house shape
%2) Character/social entity of the box and robot %audio, lights (remember that the electronic parts were to give social character to box
%(3) Perception/experiences (how 1,2 affected the experience) of the overall unboxing experience %exciting, interesting, connected, more social

\paragraph{The Appearance of the Box}
%exterior, interior, components,
From the study, children explained how the aesthetics of the box (\textit{i.e.,} exterior design, interior design, box components) affected their perceptions of the overall experience with the robot. Children showed high interest toward the \textit{physical shape and exterior design} of the box, specifically pointing out the shape of the house, magnet opening, windows, and the integrated charging outlet. Children felt that the house shape gave the robot a designated spot in which to play and live, making the overall experience more realistic. C15 stated, \textit{``I think it's the home [shape] that makes Miko feel more real and like a neighbor!''} Children also explained that the details of the house (\textit{e.g.,} roof, bricks, window, and door) increased their excitement about meeting the robot and opening the box. Also, the ease of opening the box (\textit{i.e.,} the magnetic opening) was recognized by many children as a design feature that would facilitate their interaction with the robot. Furthermore, children positively reacted toward the \textit{interior design and components of the box}. They liked the wallpaper and interior components reflecting the robot's theme, the robot’s decorative bed, and the memos on the wall. Children expressed that the box reflecting the theme through the wallpapers was exciting, and that it clarified what to expect or do in upcoming activities. When all children compared this designed box to an ordinary, regular box, they explained that this box is much more decorative and fun, making them want to keep the box as a place for the robot to stay.

\paragraph{Social Aspects within the Unboxing Experience}

Throughout the study, children discussed various design features during the unboxing that affected their social experiences. First, children responded to the idea of the \textit{box having a social entity and a purposeful character} favorably. Children described that the social presence of the box strengthened the social aspect of the robot, making the robot seem more friendly and approachable. One child (C15) stated, \textit{``He feels like a friend itself, but the house just makes it feel more like a friend. Because the house is Freddy.''} Another child (C12) suggested that the robot introduction provided by the box made the robot seem \textit{``happy to meet me and play with [me].''} The character of the box was also appreciated by children, as the role of the box being a butler for the robot not only amused children, but it transformed the role that the packaging played within the unboxing experience. Comparing to C11’s personal experience of unboxing consumer robots, C11 stated \textit{``It [the box] wasn't the main part but I feel like it is helpful and that it has a purpose now as a butler, it didn't before.''}

Although most children supported the idea of the box having a social presence, some children were ambivalent toward the robot's character. They were more interested in the supportive role of the box, particularly how it assisted the robot and provided instructions. When asked about how they felt about the robot's character, one child (C14) pointed out an inconsistency in the design of the character, stating \textit{``I liked it, but it's kind of unusual because the butlers usually keep the house, keep it clean and stuff, not be the house.''} When questioned about the box and its impact, another child (C11) described how the box was simply a secondary feature to the robot, stating \textit{``It [the box] didn't have a negative or positive effect on the robot. I feel it was just something else there in the background that helped the robot.''}

Additionally, children showed high interest in the \textit{visual and audio components of the box}, indicating that they sparked curiosity, served as a form of communication, and supported inclusiveness. Children explained that opening the box to see the lights and hear Freddy's voice was highly attractive, making them curious and excited about what comes next. One child (C16) described this experience as follows: \textit{``it made me like surprised, I didn't know that Miko would bring his house and that his house can do something, like his house is a robot.''} When asked about how they felt about this surprise, C16 stated \textit{``[in a] good way, like I didn't know that all this cool stuff was going to happen. But now I know I'm like `wow this is gonna be more exciting than I thought it was!' ''} Using its visual and audio features, Freddy sparked curiosity and enthusiasm in children and provided encouragement through the first interaction.

The visual and audio features also acted as interactive communication cues to children, as some children mentioned that Freddy's lights helped the robot express emotion and Freddy's voice resembled human behavior. Children explained that the lights blinking with various colors when they first opened conveyed the emotional state of the robot. For example, S4 stated, \textit{``The box blinking the lights made me feel the robot was happy,''}, and suggested that other emotions of the robot can be shared through the light colors in future designs. Children also explained that the audio had a social effect, as the it resembled human speech, which led them to feel like interacting with a human or a companion. When asked about their perceptions of the box verbally delivering instructions, most children compared their experience to reading a traditional instruction manual. They appreciated that that they did not have to go through instruction manuals, which are often difficult and boring to read, and suggested that Freddy explained the instructions better, which made the activities more fun and easy. One child (C11) commented on how the visual and audio features promoted inclusiveness. C11 explained how these features can ease the tension of first meeting a robot as shown in the exchange below:
\begin{quoting}
\textbf{Experimenter:} What did you think about the lights and audio?

\textbf{C11}: I think the lights and audio were a good way to represent it more as a robot yet make it more comforting.

\textbf{Experimenter:}  By robot do you mean the house or Miko?

\textbf{C11}:  House, because some people I feel like might feel a little bit distant to robots and think of them as something else, but I think people could find this more inclusive, to give it lights and audio that make it feel closer to you.

\textbf{Experimenter:}  You mentioned inclusive, why would that be?

\textbf{C11}:  I feel like some people wouldn't want to experience and interact with [the robot], but I feel like more people would like to use it because of the audio and the lights.
\end{quoting}

Finally, children expressed strong excitement towards the \textit{background story} included in the interaction design, which connected the box and the robot. The context and character given through the background story was appreciated, as it made the robot feel more like a companion. One child (S4) specifically described that the robot having a story made the robot seem \textit{``nice,''} and Freddy explaining it made both the box and the robot seem \textit{``friendly.''} Another child (C13) described this experience as follows: \textit{``[it is] cool that he had a butler [and] neat that he was a retired social robot. Made me feel that he had a friend and his own life. I like that there is a background story, it's not just here as a robot.''}

%Children also explained that the box made the overall experience more exciting, interesting, and curious compared to before. One child explained the reason for this to be \textit{``\textbf{C17}:  because adults and machines actually used their imagination to build like kids do.''}

\section{Discussion}

Our observations of children’s initial interactions with the social robot revealed a number of common factors that shape the experience of unboxing a social robot. Children were visually attracted to both the exterior and interior design of the box, highlighting the box as an opportunity for design to offer children a more positive unboxing experience that positively shapes initial interactions with the robot. Children proposed several unique solutions to design problems, such as supporting the robot with additional context, connecting the box to the robot’s story, and re-designing the box for reusability. For example, children often described the box as a place for the robot to return to after play and imagined designs that would connect the box to the robot (\textit{e.g.,} a house, a spaceship). These imagined stories helped children form common ground with the robot, broadening the ways in which they could think of interacting with the robot. Children perceived the first interaction with the robot to be similar to first meetings with peers, suggesting icebreakers and tutorial-type activities to facilitate initial interactions. 
%Functionality was also a concept that emerged during co-design, as children emphasized that the box had to be easy to open and carry and to allow taking the robot out easily. 
Overall, children did not feel limited to traditional robot technology. They suggested new ways of designing conventional materials included in technology products, such as instruction manuals, and expressed a desire toward more interactive ways of the robot's initial set up.
%Children perceived the first interaction robot similar to first meeting peers, suggesting icebreaker activities or tutorial level activities for first interaction activities.
% These findings were further supported in the co-design session.
%design spaces where imaginative stories, creative details, and novel design methods can be applied.

From the co-design study, our findings indicate that children not only perceived the robot as a social entity, but they also considered how the various entities (\textit{i.e.,} the robot, the box, and children) interacted. As a result, the unboxing experiences that children envisioned covered a broader scope than just the robot, using meaningful designs for additional narratives and social behaviors. Within their designs, children connected the box to the robot and shared narratives that \textit{bonded the robot and the box socially}. Design elements related to form and aesthetics were added to support this narrative, such as the shape of the box, exterior/interior design of the box, and box components.  %reflect the robot’s purpose, story, and capabilities. 
% Children also added personalized decorations that they thought would fit the robot, supporting the child’s bond with the robot. 
The social activities children designed emphasized personalization and communication during initial interactions, focusing on building and maintaining interest in and connection with the robot. These elements culminated in a narrative that connected the robot, the box, and the child.
% Using these findings we designed our own unboxing experience with a box that has a social entity. Our curated unboxing experience was composed of a cooperative story between the robot and the box, aesthetically meaningful box design with social characteristics, and interactive activities. These elements culminated in a cooperative narrative experience between the robot, the box, and the child.

Our evaluation of the designed unboxing experience indicated that key design points from the detailed aesthetics and the social aspects of the box contribute to positive perceptions of the social robot in children. %We added meaningful context to   to strengthen children's receptivity toward the robot. % and engagement with the robot activities.
The creative details and stories weaved into the design of the box helped children imagine, connect, and socially engage with the robot. The social presence of the box also helped children form a connection with the box, which resulted in strengthening the child's companionship toward the robot. We believe that social robot designs that offer a social connection through the unboxing experience will positively contribute to children's perceptions of and relationships with social robots over the long term.
% As key design points, we filled the aesthetic design spaces with meaningful context, which strengthened children's receptivity toward the robot and engagement with the robot activities. The creative details and stories weaved into the design of the box and initial introduction helped children imagine, connect, and socially engage with the robot.As children formed a social connection to the box and its character, this connection resulted in further strengthening the child's perception of companionship towards the robot.%Children noticed and appreciated these design details, and expressed increased connection towards the robot and engagement to the overall experience. 
% From this experience, children expressed increased excitement, connection, receptivity, and positive regard toward the social robot. 
% One child explained the reason for the positive experience to be \textit{``\textbf{C17}:  because adults and machines actually used their imagination to build like kids do.''} We also believe that the positive perceptions described above resulted as our design choices were reflective of the children’s perspectives. 

These studies provide us with valuable insights into children's experiences of unboxing a social robot. 
% More importantly, we deeply explored it from the children’s perspective which was noticed and appreciated by them. %, as one child described \textit{``\textbf{C17}: adults and machines actually used their imagination to build like kids do.''} 
We believe that the initial interaction with a social robot starts from the very moment it is delivered to the child and holds great potential for improving how a social robot is perceived by the child. Rather than being presented as another technical device in children's homes, offering a socially interactive unboxing experience from the very moment the child interacts with the box can allow for the robot to naturally merge into the child's life through connection-building. This experience can contribute to an enhanced child-robot relationship and overall positive experience with robotic technologies. 
% The results of our study highlight that a successful unboxing experience can improve children’s interest in and connection with social robots.
% As children expressed increased connection towards the robot from the unboxing experience, this positive experience can contribute to an enhanced robot experience overall. 

\subsection{Design Implications}
Throughout the course of our studies, we uncovered multiple important design solutions to support children's unboxing experience of a social robot. These design solutions integrate aesthetic creativity, making the box reusable, designing the box with social presence, developing a storyline involving the robot and the box, and having interactive activities with the robot. The high-level design implications of this work center around the four stages of the unboxing experience, \textit{i.e.,} prior interaction, packaging, first interaction, and first impression, which can anchor design decisions given specific use cases and target populations. We further elaborate on the design implications of these findings below.

\paragraph{(1) Integrate social behavior into box design.} We recommend exploring ways of integrating social behavior in the design of traditional packaging materials (\textit{e.g.,} box and instruction manuals). In the unboxing experience we designed, the instructions were delivered by the box character, \textit{Freddy}, in a social manner. This approach can transform the mechanical procedure of unboxing and setting up the robot to an interactive process that is guided by the experience design. The guided social interaction can also contribute to the development of rapport between the child and the robot toward establishing a long-term relationship. 
% This would allow the user to create a bond and build a positive first impression about the robot and overall experience. 

\paragraph{(2) Support aesthetic experience.} Our second recommendation is to offer the child an aesthetic experience through the exterior and interior design of the box as well as interactive behaviors, including light displays, color, sound, and characters. Our design of the box, based on the findings from the co-design studies with children, followed the metaphor of a ``house'' where the robot can eat, sleep, and stay. The visual, material, and interactive design of the interior supported the activities that the robot would support and thus reflected the robot's purpose of visiting the child. %the designs can be used to communicate messages, help interaction, make children interested

\paragraph{(3) Develop a backstory.} A key feature of our design was the connection between the box and the robot through a \textit{backstory}, which not only brings the robot and the box together, but it reinforces the narrative for the robot's interaction with the child, as also highlighted in prior work \cite{Simmons1111}.

\paragraph{(4) Engage the child through interactive activities.} A carefully curated unboxing experience creates opportunities to guide children's initial interactions, and our recommendation is to develop interactive activities that can serve as icebreakers to facilitate these interactions. In our design, the activities were simple tutorials, allowing children to understand the capabilities of the robot, including the physical way in which to interact with the robot, the types of activities it can perform, and the manner in which to communicate and interact with it. These activities can also serve as a form of handoff from packaging to the robot to initiate and guide child-robot interaction.

\subsection{Limitations \& Future Work}
Our work has a number of limitations. First, the exploratory and qualitative nature of our study limits our ability to offer generalizable findings. Second, logistical challenges, such as the need to drop off study resources at participant homes, conduct studies online, and pick up the resources, limited the size and diversity of our sample population. However, the co-design methods we used and developed and the insights gathered from our design studies can be applied to a larger and more diverse population of users to arrive at more generalizable and conclusive findings. Finally, although the evaluation of the designed prototype was conducted at homes in familiar environments to children rather than a lab setting, it was still a short term, one-time evaluation and deployment. This one-time interaction of the unboxing experience might not reflect how this experience will carry over to a longer-term engagement with social robots in home environments. Future work should study real-world unboxing experiences involving products users purchased and intend to use. It should also compare the effects of different designed unboxing experiences on long-term child-robot interactions and relationships by providing participants with boxes and robots that they can keep. Low-cost fabrication methods or collaboration with manufacturers of social robot products for children can enable such realistic explorations.
% Secondly, ecological validity of the design components may be of question. For example, the social entity of the box being a butler might not be valid for many social contexts \todo{such as?}. However, we have used the social entity of the designed box to create a narrative, and can be adjusted \todo{?????}

\section{Conclusion}
In this paper, we studied children's experiences with \textit{unboxing} a social robot, designed a new unboxing experience \textit{for} and \textit{with} children through participatory design, and evaluated how this experience influenced children's perspective of social robots. %evaluated this with a final free-play activity. 
In our studies, we partnered with children to understand the social bonds between children and robot technology. Our designed unboxing experience included a box conveying a social character, a four-phase carefully curated unboxing experience, and multiple themes for the social robot. Our evaluation showed that the unboxing experience of social robots can be improved through the design of a creative aesthetic experience that engages the child socially to guide initial interactions and positively structure children's relationships with robots. Our work offers several design implications for social robot products that are designed for children.

% We presented empirical evidence characterizing children's unboxing experience with . Moreover we discussed how this initial experience plays a key role in positively structuring children's perception towards social robots.
%mention that our findings address the existing pitfalls/misunderstandings between user expectation and taken-for-granted user purposes (make sure to briefly mention in intro.) Although we are not narrowly concluding this as our only finding (as we are exploring and suggesting/presenting a design space worth of exploring, it can be a meaningful contribution 
 
% \section{Appendix}
% \subsection{Key Components and Design Prompts}
% \subsection{Design Challenge}
% \subsection{Introduction Letter}
% \balance{}
% % BALANCE COLUMNS
% \balance{}

%\section*{Acknowledgements}
\begin{acks}
We would like to thank the families who participated in our studies and Dakota Sullivan for his help with the implementation of our box prototype. This work was supported by the National Science Foundation (under DRL award \# 1906854) and the University of Wisconsin--Madison Department of Computer Sciences (through the Summer Research Assistant Program). 
\end{acks}
\balance
\bibliographystyle{ACM-Reference-Format}
\bibliography{bibliography}

\end{document}